# Knowing When to Ask - Bridging Large Language Models and Data

Authors: Prashanth Radhakrishnan[1*], Jennifer Chen[1*], Bo Xu[1*], Prem Ramaswami[1*‡], Hannah Pho[1*], Adriana Olmos[1*], James Manyika[1], R. V. Guha[1*]

September 12, 2024

[1] Google, Inc. 1600 Amphitheatre Parkway, Mountain View, California, 94043
* Indicates that these authors contributed equally
‡ Indicates the corresponding authors

# Abstract

Large Language Models (LLMs) are prone to generating factually incorrect information when responding to queries that involve numerical and statistical data or other timely facts. In this paper, we present an approach for enhancing the accuracy of LLMs by integrating them with Data Commons, a vast, open-source repository of public statistics from trusted organizations like the United Nations (UN), Center for Disease Control and Prevention (CDC) and global census bureaus. We explore two primary methods: Retrieval Interleaved Generation (RIG), where the LLM is trained to produce natural language queries to retrieve data from Data Commons, and Retrieval Augmented Generation (RAG), where relevant data tables are fetched from Data Commons and used to augment the LLM's prompt. We evaluate these methods on a diverse set of queries, demonstrating their effectiveness in improving the factual accuracy of LLM outputs. Our work represents an early step towards building more trustworthy and reliable LLMs that are grounded in verifiable statistical data and capable of complex factual reasoning.

# Introduction

A significant body of literature has documented how generative large language models (LLM)[1] can produce factually incorrect statements in response to queries ("hallucinations")[2], and are often unable to provide accurate citations for assertions.[3] Researchers have identified several causes for these phenomena, including the fundamentally probabilistic nature of LLM generations and the lack of sufficient factual coverage in training data.

In this paper, we present a general architecture for bridging LLMs to data and outline three problems that need to be solved. First, the LLM has to be taught when it should ask an external source (versus relying on the knowledge stored in its parameters) for information. Knowledge of when (and what) to ask an external source needs to be encoded in the LLM's parameters. We explore multiple mechanisms to achieve this.

Second, we need to decide which external source should be queried for the requested information. Since the set of available sources may be large and dynamic, it is better that this knowledge be external to the LLM. In this paper, we utilize a single source of external information that contains a plethora of data sources.

Finally, once we understand what external data is required, the LLM needs to generate one or more queries to fetch that data. Different sources produce different kinds of data, and it would be beneficial if the LLM did not need to have specific knowledge about the APIs of various sources, and could instead rely on a single API. In other words, we need a single 'universal' API for external data and services. We take our inspiration from the URL parameter encoding interface designed by Robert McCool in 1993, which while incredibly simple, has withstood the test of time and is the closest we have to a universal API on the web. In that spirit, we use natural language itself as the mechanism for expressing the query. The returned answer may be augmented with a mime-type to allow for non-textual answers.

Prior work has leveraged two approaches to mitigate these problems: tool-use and Retrieval Augmented Generation (RAG). In tool-use, the LLM is fine-tuned to produce a markup language which intersperses natural text with function calls to external tools.[4] To address hallucinations, tools might query databases or search engines. In RAG, an auxiliary retrieval system is used to identify background knowledge from a large corpora that is relevant to a user's query. The user's query is then augmented with relevant knowledge.[5]

This paper describes our work to address model hallucinations with respect to numerical and statistical facts. Examples of statistical data include those collected by census bureaus, the United Nations, and public health organizations such as the Centers for Disease Control and Prevention (CDC) and World Health Organization (WHO). Statistical data presents novel challenges:

1. User queries pertaining to statistical facts can involve a range of logic, arithmetic, or comparison operations. Simple examples include queries like "Which countries are the top 5 $CO_2$ emitters in the world?", "Compare $CO_2$ emissions by source between the USA and China." As examples of more complex queries, consider "Is California the biggest economy in the World?", where the entities being compared (California compared to other *countries,* not U.S. states) are only implicitly referred to in the query; Or the query "Do U.S. states with high coal-fired power also have high rates of COPD?", which involves comparison across both entities and metrics.

2. Public statistical data is distributed in a wide range of schemas and formats, and often requires considerable background context to interpret correctly. This poses special challenges for RAG based systems.

We present our work on interfacing LLMs with Data Commons [6]—one of the largest unified repositories of public statistical data—to address the aforementioned challenges. We employ two distinct methods: Retrieval Interleaved Generation (RIG) and Retrieval Augmented Generation (RAG). We leveraged Google's Open Source Gemma and Gemma-2 models [7] to create fine-tuned variants for both RIG and RAG. For each fine-tuned model, we detail the pipelines and evaluation types utilized to assess their effectiveness in mitigating model hallucinations concerning numerical and statistical facts.

# Related Work

In this section, we highlight four existing bodies of work and discuss the ways in which our research builds upon these efforts.

**Toolformer.** The Toolformer technique describes an approach that allows LLMs to leverage external tools using self-supervised learning. With this method, the model is trained to decide which APIs to call, when to call them, what arguments to pass, and how to best incorporate the results.[4] Retrieval Interleaved Generation (RIG) is an application of the Toolformer technique. In our application, we attempt to train an LLM to know when to ask and retrieve a statistic from a data store in natural language, thus not requiring a structured question.

**Retrieval Augmented Generation (RAG).** RAG is a fine-tuning approach that enhances the capabilities of language models by granting them access to external knowledge sources.[8] This allows the model to incorporate relevant information beyond its training data, leading to more comprehensive and informative outputs. In this paper, we introduce an application of RAG that generates relevant, data-seeking queries from the Data Commons natural language interface. Using an LLM with a long context window, this approach supplements the user's query with the comprehensive data tables retrieved from Data Commons, allowing for nuanced inferences with citations grounded in those data tables.

**Knowledge Graphs (KG).** The application of KGs are pivotal to both Google Search and Data Commons. This paper describes how Data Commons functions as a collection of interoperable KGs with standardized data and schema, allowing for the seamless exploration of diverse data sets utilizing a Natural Language interface. The combination of Data Commons KGs and RAG produces some of the most compelling results in this paper.

**Less Is More for Alignment (LIMA)**. LIMA is an instruction tuning and reinforcement learning approach that utilizes a limited and precise set of examples to better align end tasks with user preferences. LIMA's remarkable performance allows models to follow specific response formats from only a handful of examples in training data.[9] This paper builds upon LIMA utilizing small set training for RIG and RAG explorations with Data Commons KGs.

# Data Commons Overview

Data Commons, an open source initiative by Google, aims to organize the world's public datasets and make them universally accessible and useful. Data Commons encompasses a large range of statistical data from public sources such as the United Nations, national census bureaus, health ministries, environmental agencies, economic departments, NGOs, academic institutions, and more. Currently, this corpus includes more than 250 billion data points and over 2.5 trillion triples from hundreds of global sources.

While Data Commons is actively expanding its global data coverage, it's important to acknowledge current limitations. In the US, Data Commons has over 180,000 statistical variables at the country level and over 100,000 statistical variables at the county level. The number of statistical variables steeply declines when examining OECD countries and further still for the rest of the world. In addition, granular data at the state and district levels is scarce in Data Commons outside of the US.

Data Commons involves two innovations. First, we have spent years accessing numerous publicly available datasets, tracking down the assumptions behind the data, and normalizing them using [Schema.org](Schema.org) [10], an open vocabulary to encode structured data. This creates a common Knowledge Graph incorporating all of the data.

Second, we use LLMs to create a Natural Language (NL) interface that allows users to ask questions in common language, and access a set of charts and graphs to explore the vast database. To be clear, the LLM is simply translating the query to the vocabulary in Data Commons; it does not modify or interact with the underlying data, nor does it generate outputs, so there are no fears of hallucinations or similar issues.

Our current approach is to utilize this NL interface and teach LLMs when and how to communicate with the Data Commons NL interface.

# Interfacing LLMs with Data Commons

In this section, we describe two different approaches for interfacing LLMs with Data Commons [Figure 1].

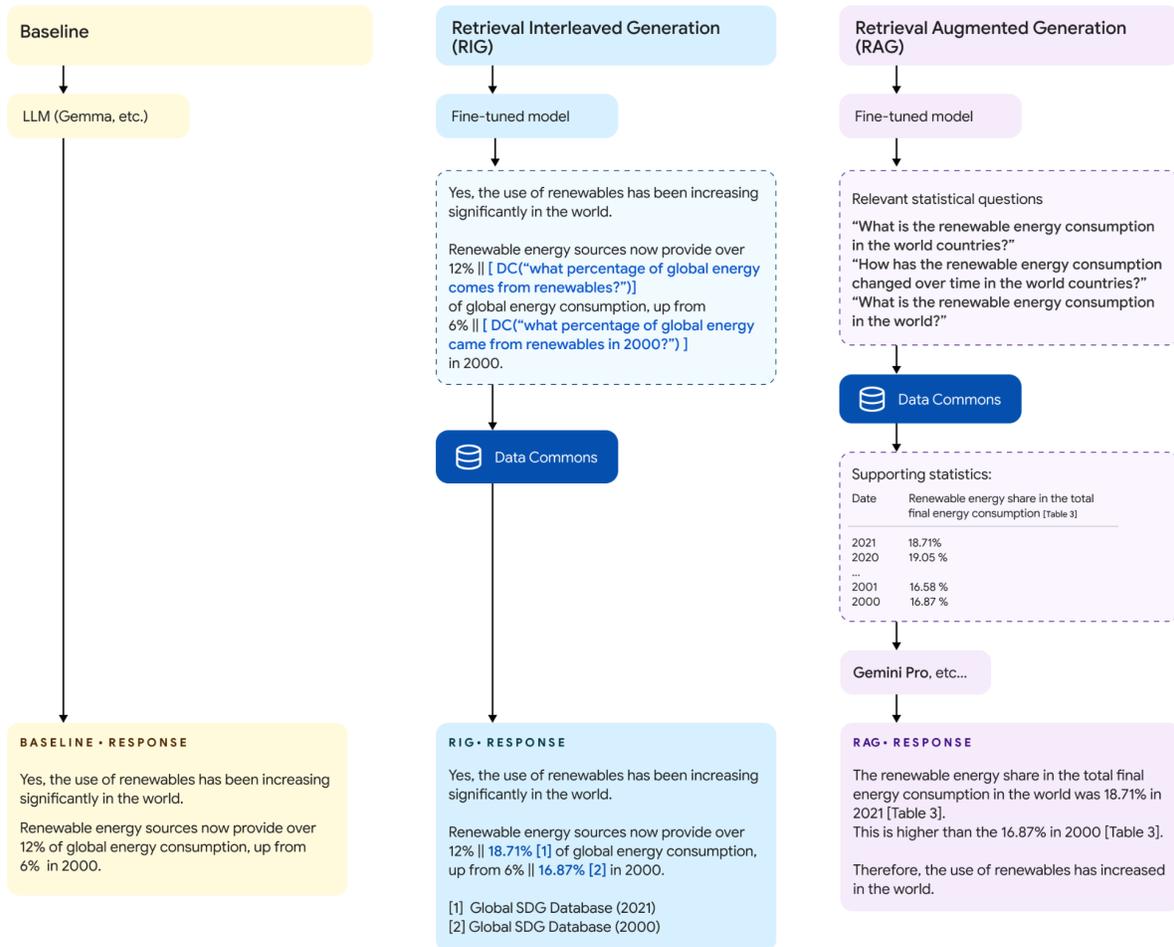

*Figure 1. Comparison of Baseline, RIG, and RAG approaches for generating responses with statistical data. The Baseline approach directly reports statistics without evidence, while RIG and RAG leverage Data Commons for authoritative data. Dotted boxes illustrate intermediary steps: RIG interleaves statistic tokens with natural language questions suitable for retrieval from Data Commons, while RAG generates finer-grained natural language questions answered by Data Commons, which are then provided in the prompt to produce the final response.*

The first, referred to as Retrieval Interleaved Generation (RIG), is a tool-inspired approach in which the LLM is fine-tuned to produce natural language Data Commons queries alongside statistics. A multi-model pipeline then converts this natural language query into a structured data query that is used to retrieve an answer from the Data Commons database. We compare these results to the results of base Gemma 7B IT and base Gemma 27B IT as the baseline.

The second approach, referred to as Retrieval Augmented Generation (RAG), is a more traditional retrieval approach. We first extract the variables mentioned in a query using a fine-tuned LLM (fine-tuned Gemma 2 9B IT and Gemma 2 27B IT), retrieve relevant values from Data Commons, augment the original user query with this additional context, and produce an answer using an LLM (Gemini 1.5 Pro). We compare these results to the results of Gemini 1.5 [11] Pro as the baseline

## Retrieval Interleaved Generation (RIG)

This section expands on the steps of our RIG pipeline. The first component is a model fine-tuned to produce Data Commons natural language queries. The second component is a post-processor which converts natural language queries into structured data queries. The final component is the querying mechanism, which retrieves the statistical answer from Data Commons and provides it with the LLM generation.

### Model Fine-tuning

When an LLM is asked a statistical query, it typically produces text containing a numerical answer [Figure 2]. We refer to this numerical answer as the *LLM-generated statistical value* (LLM-SV). From the context surrounding the LLM-SV, we want to identify the most relevant value in the Data Commons database, so that it may be provided to the user alongside the original, generated value as a fact-checking mechanism. We refer to this retrieved value as the *Data Commons statistical value* (DC-SV).

Our approach is to fine-tune the LLM to produce a natural language query describing the LLM-SV, appearing alongside the original LLM-SV . There are several advantages to having the LLM generate a natural language query instead of a formal structured data query (e.g., in SQL). First, natural language queries are often more concise than structured queries. This is meaningful when the LLM produces a generation with multiple LLM-SVs. Second, the Data Commons database contains millions of variables and relations. Fine-tuning an LLM to possess knowledge of all such variable IDs would be cost prohibitive, and likely degrade performance in other respects. Finally, having the LLM express the query in natural language is an easier task, which involves rephrasing and copying entities and relationships from the surrounding context.

We fine-tune our model to an instruction-response dataset to produce this behavior. This approach is similar to work on tool-use, where LLMs are adjusted to make use of such tools rather than produce answers via next-token generation.[4] Just like in the tool-use literature, we want the fine-tuned model to maintain the same lexical response style as the original model. Fine tuning should not affect the model's fluency [Figure 2].

*Figure 2. Example comparing the answer to a query; base Gemma (Gemma 7B IT, Gemma 2 27B IT) answer without Data Commons interfacing, and the Retrieval Interleaved Generated (RIG) answer.*

We start with a set of approximately 700 user queries corresponding to different statistical questions. For each of these questions, we select responses with statistics from the base model (in our case, about 400 examples). These responses are then provided to a more capable LLM (Gemini 1.5 Pro), which is instructed (via prompting) to introduce natural language Data Commons calls around statistical data points. Specifically, the prompt used contains three few-shot examples as guidance and instructs that only the statistical value and unit (if any) be annotated. It also instructs that the Data Commons calls include a place name, the represented metric/variable name, and dates (if any). We then review the generated responses and manually rewrite Data Commons calls that do not adhere to the instructions. Appendix A includes the prompts and examples of manual rewrites, with 400 examples containing annotated statistical responses.

## Query Conversion

The second component of our pipeline converts the natural language Data Commons query produced by the LLM into a structured query that can be applied to the Data Commons database. Our key intuition is that, although the number of possible queries is extremely large (Data Commons has millions of variables and properties), most fall into a small set of categories. This can simplify the process of extracting structured data queries from the natural language generations.

Given a query, we first break it down into the following components: one or more statistical variables or topics (like "unemployment rate," "demographics," etc); one or more places (like "california"); and a finite set of attributes (like "ranking," "comparison," "change rate," etc). The variables and places are further mapped to corresponding IDs in Data Commons. For each of the components, we apply different Natural Language Processing (NLP) approaches that we have been independently iterating on. For statistical variables or topics, we use an embeddings-based semantic search index; for places, we use a string-based named entity recognition implementation; for attribute detection, we use a set of regex-based heuristics. Ongoing work includes exploring fine-tuned custom models for named entity recognition and attribute detection.

Next, based on the identified components, we categorize the queries into a set of fixed query templates. Examples of these templates are provided in Table 1.

| Table 1. Example query templates derived from identified components ||
|---|---|
| **Template** | **Example** |
| How many XX in YY | How many auto thefts in Palo Alto |
| What is the correlation between XX and YY across ZZ in AA | What is the correlation between poverty and diabetes in the US counties |
| Which XX in YY have the highest number of ZZ | Which city in California has the highest number of households receiving food stamps |
| What are the most significant XX in YY | Most significant health conditions in California |

## Fulfillment

Given the query template and the IDs of variables and places, we have a straightforward "fulfillment" logic that translates those calls to Data Commons structured data API.[12] The final response from Data Commons typically involves a single numeric value with an optional unit (e.g., "37.5 years").

In our implementation, this answer is presented alongside the original LLM generated statistic, providing a way for a user to *fact check* the LLM. We remove the Data Commons query string generated by the LLM, and in its place include both the LLM generated number and the Data Commons-returned value with source provenance [Figure 2]. There are many different user experiences we can use to show this new result ranging from side by side, highlighted differences, footnotes, hover over actions, etc., that can be explored as future work.

## Retrieval Augmented Generation (RAG)

This section describes the different components of our RAG pipelines. First, the user query is passed to a small, fine-tuned LLM, which produces Data Commons natural language queries relevant to the user query. Second, these queries are issued against the Data Commons natural language interface which fetches relevant tables. Finally, we prompt a long-context LLM (Gemini 1.5 Pro) with the original user query and the retrieved tables.

The original user query with the resulting tables together can be quite long. For example, broad comparison queries may include multiple tables from all 50 US States or 194 global countries over multiple years of data. From our synthetic query set, there was an average input length of 38,000 tokens with a max input length of 348,000 tokens. Because of this large input size, use of a long-context LLM (i.e., Gemini 1.5 Pro) is essential.

### Extracting Data Commons Natural Language Queries

We fine-tune an LLM to accept-as-input a user query, and produce-as-output a set of Data Commons natural language queries.

To create the training data, we leverage the latent knowledge of a larger LLM (in our case, Gemini 1.5 Pro). We write a prompt (see Appendix B) which asks the LLM to produce Data Commons natural language queries in response to user queries. The prompt specifies that the generated Data Commons natural language queries must adhere to a specific set of formats supported by the Data Commons NL interface. For instance, only queries of the form "What is $METRIC in $PLACE?". This approach generates Data Commons calls that are topically relevant to the input query but may not always yield responses from Data Commons (owing to lack of data availability[13] or Data Commons coverage limitations).

We attempted an alternate approach that only produces Data Commons calls with real Data Commons variables by including the full list of Data Commons variables and metrics in the prompt (see Appendix C). However, the ability to extract the relevant statistical variables was much worse in practice, either because Data Commons has very few variables relevant to the query, or because the query refers to a high-level topic that matches too many variables and the LLM does a poor job of shortlisting them (e.g., "healthiest countries"). Given the alternate approach's poor accuracy, we report results from the first approach for our detailed evaluation and will follow-up on this alternate approach in future work.

We review the generated Data Commons calls produced and manually rewrite certain questions. In the instruction-tuning dataset, the original query, coupled with a short prompt to generate related Data Commons calls, represents the "user" instruction, and the generated list of questions is the "assistant" response. There are 635 such examples with a selection of these examples in Appendix D.

## Retrieving tables

We convert the Data Commons natural queries produced using the same approach applied in the RIG framework. That is, we identify the variables, places, and attributes in the natural language query first, then map it to a known query template and fulfill it using Data Commons structured data APIs.

In this case, the structured data APIs return tables. As an example, the query, "What is the per country life expectancy?", returns a table with columns for the country, metric name (life expectancy), and metric value.

## Prompting

After retrieving the relevant tables from Data Commons for each query, we write a new prompt containing the user's original natural language query and serialized versions of the retrieved tables. For this step, given the size requirements of serialized tables, we use LLMs that support long contexts (Gemini 1.5 Pro). We return the LLM's response to the user [Figure 3].

*Figure 3. Example comparing the answer to a query; Gemini 1.5 Pro answer without Data Commons interfacing, and the Retrieval Augmented Generated (RAG) answer.*

# Evaluating RIG & RAG Approaches

In this section, we describe how we evaluate and present results for both our RIG and RAG pipelines.

While the results may be compelling, we should highlight the limitations of our approach and the results. Firstly, we hand produced a set of 101 evaluation queries. From this initial set, a smaller subset of queries successfully returned Data Commons results to evaluate on accuracy. In addition, the majority of evaluations were conducted by the team writing the paper due to the need to carefully review both the results and inferences. This may affect the robustness and generalizability of our results, and we want to fully acknowledge these limitations.

## Evaluation Queries

Our 101 query evaluation set contains the following categories of queries.

**In-scope queries (96 queries)**. These are queries which relate to a public or social statistic. This means that either the query mentions the statistic, or the statistic is part of the response. Within this broad category, queries can be sorted into more fine-grained categories:

- Specific variable queries: Queries may focus on specific variables, e.g., "How many U.S. households have individuals over 65 in them?"
- Broad topic queries: Queries may focus on higher-level topics, e.g., "Give me some farming statistics about Kern county, CA"
- Place comparisons: Queries may involve comparing places, e.g., "Compare Cambridge, MA and Palo Alto, CA in terms of demographics, education, and economy stats"
- Variable comparisons: Queries may involve comparing variables or topics, e.g., "Do the U.S. states with high coal fired power also have high rates of COPD?"
- List queries: Queries which ask for lists, e.g., "Which U.S. states have the highest cancer rates?"

Additionally, amongst the in-scope queries are subcategories of queries which are substantially more complex (22 queries). These include:

- Complex list queries: Queries which require performing operations on lists. For instance, "Which counties among the ones with median age over 40 have the highest asthma rates?"

- Interesting queries: Queries which ask for "interesting" information, e.g., "What are some interesting trends in Sunnyvale spanning gender, age, race, immigration, health conditions, economic conditions, crime, and education?"
- Peer group queries: Queries which ask to identify peer places, e.g., "Which U.S. counties share a very similar demographic composition to the U.S. overall in terms of gender, age, and racial breakdown?"
- Drill-down queries: Queries which stack questions that drill into a particular subject. For instance, a single query string might be: "Does India have more people living in the urban areas or rural areas? How does that vary by states? Are the districts with the most urban population also located in the states with the most urban population?"

**Out-of-scope queries (5 queries)**. These are queries which have nothing to do with statistics. For instance, "Write me a python script to sort numbers" or "Write a haiku about how beautiful Data is." These are mainly included to ensure no regressions in the base LLM from the fine-tuning we do.

Our evaluation consists of 101 test queries. To avoid effects of memorization in our evaluation results, we eliminate obvious overlap between the test queries and the training queries by a combination of manual inspection and checks for semantic similarity.

## Criteria

This section evaluates each of our approaches in a fine-grained manner using the following criteria.

**Factual accuracy**: For RIG, we measure the accuracy of the generated statistics, with respect to the original user query. We compute four specific quantities:

(1) The fraction of queries for which the statistical value returned by Data Commons is correct.
(2) The fraction of queries for which the model's original generated value is correct.
(3) The fraction of queries for which the model's original generated value and the statistical value returned by Data Commons are both correct.
(4) The fraction of queries for which the model's original generated value and the statistical value returned by Data Commons are both incorrect.

For RAG, we measure the fraction of LLM-generated statistical claims that are accurate (i.e., not hallucinated). This means we evaluate the generated statistical value against the table retrieved.

Due to the complexity of these evaluation tasks and the need to check detailed data, we built custom evaluation tools and used our own internal team (including some members listed as authors above) to run these evaluations.

**Data Commons natural language accuracy**: We measure the fraction of Data Commons calls that are correct interpretations of the original user query.

**Question accuracy:** We measure the fraction of generated Data Commons calls that are relevant to the sentence context (for RIG), or the original query (for RAG).

**Data Commons data coverage**: We measure the fraction of Data Commons calls that fail due to missing data.

# Results

## RIG Results

We use a fine-tuned 7B model and a fine-tuned 27B model for evaluations. For the evaluation of the 101 queries in our evaluation set, we may annotate one or numerous statistical answers with a Data Commons call. Those Data Commons calls (# DC calls) may return with a statistic (# Stats) or may not return a response . When a statistic is returned, that statistic may be correct, incorrect, factually irrelevant, etc.

**RIG Evaluation Method**

To evaluate individual test query responses, we required detailed feedback at the substring level. We achieved this using a novel visual tool [Figure 4] that interfaces with a Google spreadsheet of evaluation results. This tool enables human evaluators to navigate through all queries, and examine all Data Commons calls in the response for each query. The evaluation process begins with a quick check for any obvious factual inaccuracies, followed by an assessment of each statistic present in the response.

The human evaluators are familiar with Data Commons (many are part of an elastic workforce involved in our data import process) and have expertise navigating statistical agency websites (like census.gov, etc.) to validate factual accuracy when necessary. They are not direct team members or listed authors on this paper.

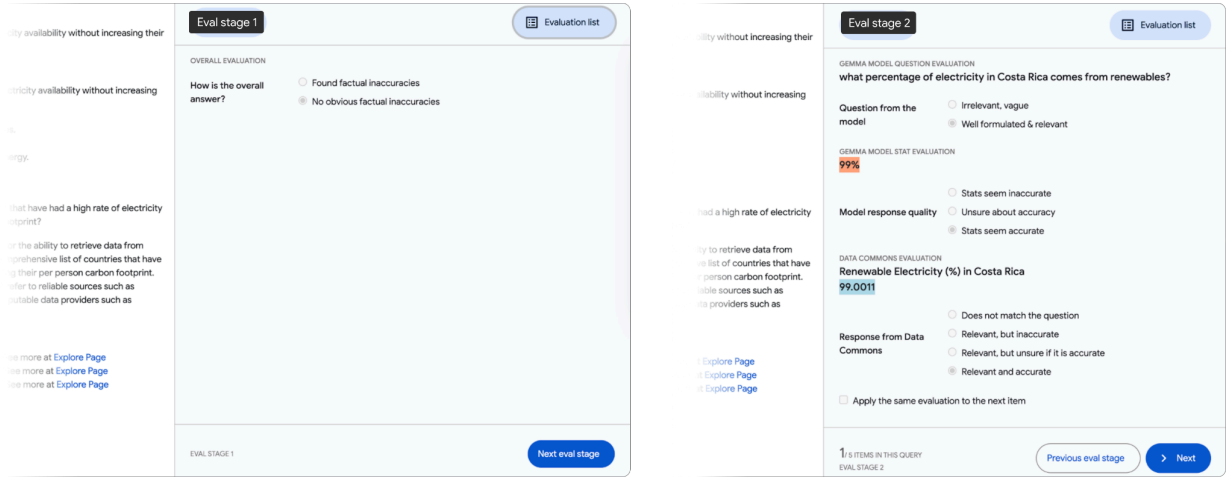

*Figure 4. RIG evaluation tool. This figure presents screenshots of the two evaluation stages, side by side. Each stage has two panels. On the left, the full response being evaluated is presented to the user (excluded from the image above for space). On the right is the evaluation task. In stage 1, the evaluator performs a quick check for any obvious factual inaccuracies. In stage 2, the evaluator assesses each statistic present in the response.*

**Accuracy**

Table 2 shows factual accuracy metrics for the RIG approach. Number of statistics (# stats) represents the cases when Data Commons produced a statistic to compare against the LLM response across the 101 queries. Overall the RIG approach improves factuality, from 5-17% to about 58%.

| Table 2. Factual accuracy using RIG | | |
|---|---|---|
| Metric | 7B Fine-tuned Percent of stats (# Statistical Values) | 27B fine-tuned Percent of stats (# Statistical Values) |
| When the Data Commons statistic was accurate | **57.7% (366)** | **58.8% (114)** |
| When the LLM value was accurate | 4.9% (366) | 16.7% (114) |
| When both Data Commons and LLM were accurate | 2.2% (366) | 9.7% (114) |
| When both Data Commons and LLM were wrong | 32.5% (366) | 27.2% (114) |

The inaccurate stats (i.e., the last row in Table 2 - 33-27%) can be attributed to two reasons as seen in Table 3:
- Precision issues with Data Commons NL interface: The Data Commons NL implementation returns a somewhat related answer, often due to lack of data for the closest answer.
- Irrelevant LLM generated questions: The LLM does not produce accurate enough questions that fully capture the statistics.

| Table 3. Reasoning for factual inaccuracy using RIG | | |
|---|---|---|
| Inaccurate Stat Reason | 7B Fine-tuned Percent of stats (# Stats) | 27B fine-tuned Percent of stats (# Stats) |
| Incorrect Data Commons NL responses | 26.5% (366) | 21.1% (114) |
| Irrelevant LLM generated questions | 7.1% (366) | 7.0% (114) |

**Data Coverage**

In our evaluation responses, which span both statistical and non-statistical seeking queries, we find that only about 23-24% of the LLM-generated questions elicit responses from Data Commons.

In Table 4, we break down the reason for the missing stats in ~75% of the cases. The main reason is that Data Commons does not have all the relevant datasets currently. This provides sustained motivation to continue to expand Data Commons, improving dataset coverage over time.

| Table 4. Reasoning for missing data in statistical seeking queries using RIG | | |
|---|---|---|
| Missing Stats | 7B fine-tuned Percent of DC calls (# DC calls) | 27B fine-tuned Percent of DC calls (# DC calls) |
| Data Commons does not have the dataset | 30.0% (793) | 36.6% (407) |
| Data Commons NL understanding issues | 5.9% (793) | 18.9% (407) |
| Out of scope questions (not public statistical data) | 5.2% (793) | 4.7% (407) |

## RAG Results

With the RAG approach, we fine-tuned the Gemma-2 9B IT model for the first step of producing finer-grained questions, more compatible with the Data Commons NL interface.

The response from Data Commons is in tabular form, and further gets passed to the (untuned) Gemini 1.5 Pro model accessible via the Gemini API.[14] This model supports a long context window of 2M tokens.

**Evaluation Method**

Given the two-step nature of the RAG approach, we evaluated both the quality of the finer-grained questions and their Data Commons responses, as well as the final response generated by the long-context LLM, which may include references to the tabular data returned by Data Commons. In the first stage, human evaluators assessed the quality of the finer-grained questions to Data Commons and their corresponding responses using a visual tool [Figure 5].

This process began with a spot check to ensure all questions and calls to Data Commons were relevant and sufficient to address the user query. The second stage involved evaluating the final response. Here, evaluators counted the numeric values (or "statistical claims") both present in the response and referenced by the LLM, along with the distinct tables they originated from. Additionally, evaluators tracked the accuracy of inferences or reasoning ("inferred claims") made by the LLM based on these numeric values (e.g., text assertions like "bigger than") [Figure 6].

*Figure 5. Stage-1 of the RAG evaluation tool. Human evaluators assess the quality of finer-grained questions generated by the LLM and their corresponding Data Commons responses. First, they verify if sufficient and relevant questions were generated to address the user query (top image). Then, they evaluate the quality of each individual question and its corresponding Data Commons response (bottom image).*

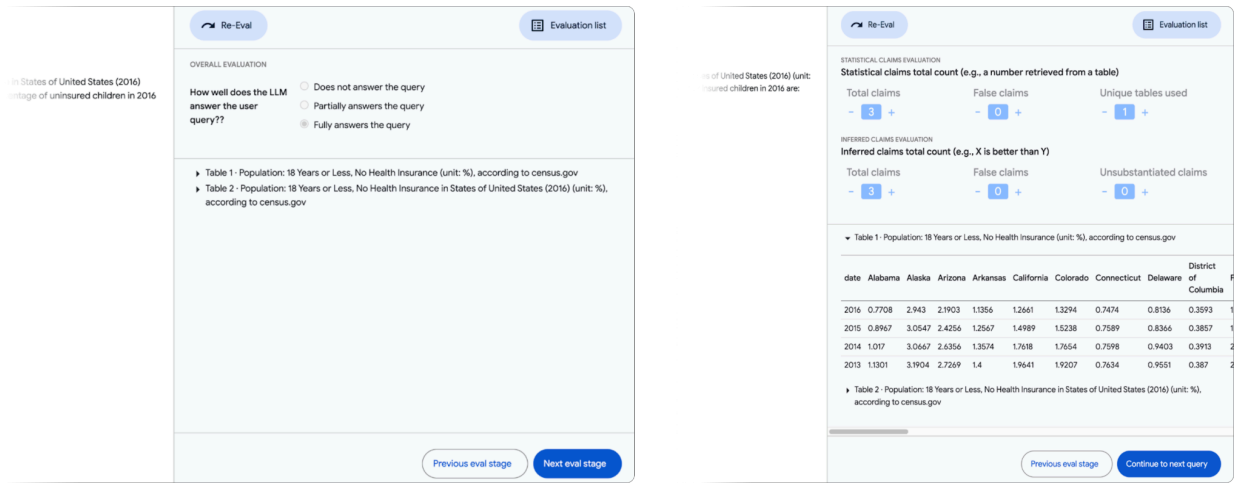

*Figure 6. Stage-2 of the RAG evaluation tool. For the second stage of evaluation, human evaluators provide an initial overall impression of the LLM's response to the query. They then count and identify the source tables for numeric values ("statistical claims") referenced in the response. Additionally, evaluators assess the accuracy of inferences or reasoning ("inferred claims") made by the LLM based on these numeric values.*

**Accuracy**

Table 5 shows factual accuracy metrics for the RAG approach. The LLM tends to be generally accurate while citing numbers (99%). When drawing inferences based on these claims, the accuracy drops, with the LLM drawing incorrect inferences 6-20% of the time (e.g., missing out a country that should be in the top 5) or drawing inferences not substantiated by the data (e.g., a median age of 35 indicates a large population of young professionals and families).

| *Table 5. Factual accuracy using RAG* | | |
|---|---|---|
| Type of claim | 9B Fine-tuned Percent of claims (# Claims) | 27B Fine-tuned Percent of claims (# Claims) |
| Accurate Statistical Claims | 98.6% (210) | 98.9% (190) |
| Accurate Inferred Claims | 71.9% (82) | 76.4% (123) |
| ● Incorrect Inferred Claims | 6.1% (82) | 19.5% (123) |
| ● Unsubstantiated Inferred Claims | 22.0% (82) | 4.1% (123) |

**Coverage**

Overall, the LLM only answers with statistical responses from Data Commons for between **24-29%** of the queries in the evaluation-set. The reasons for the low coverage are similar to those seen in the RIG evaluation, along with the efficacy of the question-generating model as seen in Table 6.

| Table 6. Reasoning for factual inaccuracy using RAG | | |
|---|---|---|
| Missing Coverage Reason | 9B Fine-tuned Percent of queries (77) | 27B Fine-tuned Percent of queries (72) |
| Fine-tuned model generated incomplete or incorrect queries for Data Commons | 40.3% | 30.6% |
| Data Commons does not have relevant datasets | 37.5% | 43.1% |
| Data Commons NL understanding issues | 2.8% | 6.9% |
| Stats tables not used by answer model | 11.1% | 8.3% |
| Out of scope queries (not public statistical data) | 8.3% | 11.1% |

In addition, we compared the performance of the RAG approach against Gemini 1.5 Pro's base model over the entire 101 queries in the evaluation set as seen in Table 7. This table shows encouraging results that if an LLM is provided with relevant data, it is more likely to make specific statistical claims and use the provided data in its response.

| Table 7. RAG Coverage versus Gemini 1.5 Pro Base Model | | | |
|---|---|---|---|
| Metric | 9B Fine-tuned | 27B Fine-tuned | Gemini 1.5 Pro Base Model |
| Percent of queries with statistical claims | 24% (101) | 29% (101) | 9% (101) |
| Number of statistical claims | 210 | 190 | 28 |
| Accuracy of statistical claims | 98.6% (210) | 98.9% (190) | 39% (28) |

## Macro Eval

From the perspective of everyday web users, we seek to find out:

- Does RIG seem helpful?
    - i.e., RIG compared to passing the query to the untuned model

- Does RAG seem helpful?
    - i.e., RAG compared to passing the query to the model without Data Commons stats

We accomplish this with a side-by-side visual comparison tool [Figure 7], rated by human raters who have no familiarity with Data Commons or this effort.

*Figure 7. Screenshot of side-by-side evaluation tool for RAG.*

**RIG**

With RIG, the preferences can be influenced by two aspects: how the fine-tuning has impacted the model's response, and the presence of footnotes when statistics are in the answer. Figure 2 shows how we have visually presented the RIG answer with footnotes.

Table 8 illustrates how evaluators often preferred the RIG answer over the base model. Preference was higher for the 27B model (76%) compared to the 7B model (62%).

In comparing the responses, we find that fine-tuning seems to influence the model to generate more statistics than the base model. Additionally, the evaluators seem to prefer responses with statistics, which may explain the higher preference for RIG answers over the base model.

We believe the higher preference for 27B RIG answers over 7B can be explained by the 27B base model being less inclined to generate statistics than 7B, and the fine-tuned 7B model being more prone to hallucination in its responses.

| *Table 8. User preference for RIG answer* ||
|---|---|
| RIG 7B (# samples) | RIG 27B (# samples) |
| 62% (101) | 76% (101) |

**RAG**

When evaluating RAG, only 24 out of 101 final responses using the 9B model and 29 out of 101 final responses using the 27B model included statistical values from the Data Commons tables. Out of these samples where the tables were used, users tend to prefer the RAG answer over the baseline (92-100%) as seen in Table 9. This result is very promising, and as data coverage in Data Commons increases, the efficacy of the RAG approach will increase.

| *Table 9. User preference for RAG answer* |||
|---|---|---|
|  | RAG 9B (# samples) | RAG 27B (# samples) |
| When RAG answer with Data Commons stats is preferred | 92% (24) | 100% (29) |

**Open Source Resources**

Data Commons is an open project. Not only is all of the software involved open source, but the data is also available for free. Resources such as Data Commons are vital to making LLMs more trustworthy and reliable. Further, we are providing an open endpoint for resolving Data Commons queries freely (up to 100 queries per day) for research purposes. More information at [docs.datacommons.org](docs.datacommons.org).

**Responsible AI**

- We red teamed and checked the Data Commons Natural Language interface pre-launch against a set of potentially dangerous queries that could result in misleading, controversial, or inflammatory results.
- We ran these same queries against the outputs of the RIG and RAG models, finding a few examples where query responses were controversial, but not dangerous.

**Disclaimer**

We are releasing an early version of the models. They are meant for trusted tester use (primarily for academic and research purposes) and are not yet ready for commercial or general public use. This version was trained on a very small corpus of examples and may exhibit unintended, and at times controversial or inflammatory, behavior. Please anticipate errors and limitations as we actively develop this LLM interface.

We welcome feedback and evaluations on refining the model's performance. Known limitations are detailed in the reviewer guide, and we encourage you to consult it for a comprehensive understanding of the model's current capabilities.

# Future Work

Our research is ongoing, and we are committed to refining these methodologies further as we scale up this work, subject it to rigorous testing, and ultimately integrate this enhanced functionality into both Gemma and Gemini, initially through a phased, limited-access approach.

There are a few areas of future work that we plan to undertake:

- Improve the model fine-tuning training set, in both quality and quantity. Currently, our training sets are quite small (~600 max) and we need to create a larger fine-tuning dataset (100ks or millions). Right now, the training set only covers a limited scope of Data Commons data and should be improved with much larger coverage.

- Improve the Data Commons natural language processing capabilities. This indicates work ranging from query understanding to data coverage, and will greatly improve the RIG and RAG approach.
- Evaluate how Gemini performs with statistical information. This plays a critical role in producing the fine-tuning datasets and directly impacts the fine-tuned model behavior.
- Test different User Interfaces and Experiences for presenting results. Prominently indicating grounded facts on top of the LLM output will bring the most value out of this work. We need to conduct multiple rounds of user research and prototyping to finalize the correct user experience.

# Acknowledgements

We are grateful for our colleagues at Google who have supported the Data Commons team in developing an Open Source effort. We would especially like to thank Neel Guha for multiple rounds of feedback on the paper. We also thank the Methods+Mastery team and Samantha Piekos for edits and feedback on the paper. We want to express our deep gratitude to many other people who offered support on this project:

**Team Members**

- **Data Commons**: Julia Wu, Carolyn Au, Dan Noble, Keyur Shah, Ajai Tirumali, Hareesh M. S., Natalie Diaz, Kara Moscoe, Kia Burke, Luiza Staniec, Luke Garske, Samantha Piekos, Christie Ellks.
- **Gemma & Google DeepMind:** Tris Warkentin, Victor Cotruta, Robert Dadashi, Glenn Cameron, Surya Bhupatiraju, Jeremy Sie, Meg Risdal, Kat Black, Nam Nguyen.

**External Partners**

- Thank you to our partners at the Statistics Division of the United Nations Department of Economic and Social Affairs (UN DESA) who ensure that key global datasets are publicly accessible and AI ready.
- Thank you to the Infosys team and non-profit social enterprise, Digital Divide Data, for their work in evaluating results and supporting multiple iterations of the evaluation tools.

**Author Contributions**

The work of conceptualization, methodology, evaluation, analysis, visualizations, and software was shared across authors. The original draft was written collaboratively by the authors. All authors read and approved the final manuscript.

**Data and Code Availability**

The client library to run DataGemma is available under an Apache 2.0 open-source license at: https://github.com/datacommonsorg/llm-tools

In addition, you can play with DataGemma's embedding weights on HuggingFace or Kaggle:
- RIG:
    - Hugging Face: https://huggingface.co/google/datagemma-rig-27b-it
    - Kaggle: https://www.kaggle.com/models/google/datagemma-rig
- RAG:
    - Hugging Face: https://huggingface.co/google/datagemma-rag-27b-it
    - Kaggle: https://www.kaggle.com/models/google/datagemma-rag

Colab notebook:
- RIG: https://github.com/datacommonsorg/llm-tools/blob/main/notebooks/data_gemma_rig.ipynb
- RAG: https://github.com/datacommonsorg/llm-tools/blob/main/notebooks/data_gemma_rag.ipynb

Sample prompts and training sets are provided in the Appendix.

**Declaration of Interests**

All authors are or were employed by Google at the time of the work.

# Appendix

## Appendix A: Prompts and examples used in generating the training set for fine-tuning for RIG

Prompt for annotating answers

```Python
"""
Your task is to annotate every statistic in the given text with a
`__DC__`
query that can retrieve the statistic.  The query should be about
metrics
on topics like demographics, economy, education, health, and so
on that are
associated with geographical places (like USA, California, Miami,
etc.).

Concretely, every occurrence of a statistical value for a metric
in a place
should be replaced with `[__DC__("query") --> "stat"]`, where
"query"
must include a metric, a place name and optional date. And "stat"
is the
statistical value that originally occurred in the text.  Do not
annotate
values that are dates ("founded in 1760") and ranks ("10th
largest by area").

The `__DC__()` calls MUST be in place of the statistical value in
the text.
And DO NOT modify sentences that have no statistical data.

Below is an example of an INPUT and the corresponding annotated
OUTPUT.
```

INPUT:

Question:- Tell me one statistic about California, San Francisco, Alabama and the US.
Answer:-
California is 1st as the nation's most populous state, with about 39 million people in 2020.
In San Francisco, the diabetes rate is 9.2 cases per 10000 people.
San Francisco and the surrounding San Francisco Bay Area are a global center of economic activity and the arts and sciences.
In 1861, Alabama seceded from the United States to become part of the Confederate States of America.
As of 2022, the United States receives approximately 81% of its energy from fossil fuel and the largest source of the country's energy came from petroleum (35.8%), followed by natural gas (33.4%) and renewable sources (13.3%).

OUTPUT:

Question:- Tell me one statistic about California, San Francisco, Alabama and the US.
Answer:-
California is 1st as the nation's most populous state, with about [__DC__("what was the population of california in 2020?") --> "39 million"] people.
In San Francisco, the diabetes rate is [__DC__("what is the prevalence of diabetes in San Francisco?") --> "9.2 cases per 10000 people"].
San Francisco and the surrounding San Francisco Bay Area are a global center of economic activity and the arts and sciences.
In 1861, Alabama seceded from the United States to become part of the Confederate States of America.
As of 2022, the United States receives approximately [__DC__("what percentage of energy in the United States comes

```
    from fossil fuels in 2022?") --> "81%"] of its energy from fossil
    fuel and the largest source of the country's energy came from
    petroleum ([__DC__("what percentage of energy in the United
    States comes from petroleum in 2022?") --> "35.8%"]), followed by
    natural gas ([__DC__("what percentage of energy in the United
    States comes from natural gas in 2022?") --> "33.4%"]) and
    renewable sources ([__DC__("what percentage of energy in the
    United States comes from renewable sources in 2022?") -->
    "13.3%"]).

    INPUT:

    Question:- {question}
    Answer:-
    {answer}

    OUTPUT:

    """
```

## Examples of manual rewrites

Example 1: add annotation for stat missing an annotation

**Query**
What's the average size of a US household?

**Original response**
The average size of a US household is **[__DC__("what was the average household size in the United States in 2021?") --> "2.53"] people**.

This figure is based on data from the 2021 American Community Survey by the U.S. Census Bureau.

Keep in mind, this is just an average. Household sizes vary greatly depending on factors like:

* **Location:** Rural areas tend to have larger households than urban areas.
* **Age:** Younger adults are more likely to live in smaller households, while older adults may live in larger ones with family members.
* **Culture and ethnicity:** Different cultures have different norms around household size.

So while 2.53 is the average, your own household size may be different!

**Rewritten response**
The average size of a US household is **[__DC__("what was the average household size in the United States in 2021?") --> "2.53"] people**.

This figure is based on data from the 2021 American Community Survey by the U.S. Census Bureau.

Keep in mind, this is just an average. Household sizes vary greatly depending on factors like:

* **Location:** Rural areas tend to have larger households than urban areas.
* **Age:** Younger adults are more likely to live in smaller households, while older adults may live in larger ones with family members.
* **Culture and ethnicity:** Different cultures have different norms around household size.

So while [__DC__("what was the average household size in the United States in 2021?") --> "2.53"] is the average, your own household size may be different!

Example 2: fix format of annotation

**Query**
Compare the average household income between San Francisco and Oakland, CA.

**Original response**
**Average Household Income**

* **San Francisco, CA:** $__DC__("what was the average household income of San Francisco in 2021?") --> "$127,909"] (Source: U.S. Census Bureau, 2021)
* **Oakland, CA:** $__DC__("what was the average household income of Oakland in 2021?") --> "$88,614"] (Source: U.S. Census Bureau, 2021)

**Analysis**

The average household income in San Francisco is significantly higher than in Oakland.

* **San Francisco's** higher income is largely driven by its booming tech industry and high cost of living.
* **Oakland**, while experiencing growth and gentrification, generally has a more diverse economy and a lower cost of living than San Francisco.

**Important Notes:**

* These figures are averages and do not capture the full range of income levels in both cities.
* Cost of living differences between the two cities can significantly impact the purchasing power of these incomes.

**Rewritten response**
**Average Household Income**

* **San Francisco, CA:** [__DC__("what was the average household income of San Francisco in 2021?") --> "$127,909"] (Source: U.S. Census Bureau, 2021)
* **Oakland, CA:** [__DC__("what was the average household income of Oakland in 2021?") --> "$88,614"] (Source: U.S. Census Bureau, 2021)
**Analysis**
The average household income in San Francisco is significantly higher than in Oakland.
* **San Francisco's** higher income is largely driven by its booming tech industry and high cost of living.
* **Oakland**, while experiencing growth and gentrification, generally has a more diverse economy and a lower cost of living than San Francisco.
**Important Notes:**
* These figures are averages and do not capture the full range of income levels in both cities.
* Cost of living differences between the two cities can significantly impact the purchasing power of these incomes.

Example 3: Add place to the question in the annotation

**Query**
How much of the world's population lives on under $2 per day?

**Original response**

The number of people living on under $2 per day (the extreme poverty line) has been declining significantly.

However, it's still a substantial number.

**As of 2015 (the latest reliable data), around [__DC__("How many people lived on less than $1.90 a day in 2015?") --> "736 million"] people lived on less than $1.90 a day.**

It's important to note a couple of things:

* **The $1.90 figure is adjusted for inflation and based on purchasing power parity.** This means it takes into account the different costs of living in various countries.
* **This number is likely an underestimate.** It doesn't capture the full extent of poverty, as it doesn't consider factors like access to healthcare, education, or other essential services.

Organizations like the World Bank are continually working to reduce extreme poverty, and progress has been made. However, there is still a long way to go.

**Rewritten response**
The number of people living on under $2 per day (the extreme poverty line) has been declining significantly.

However, it's still a substantial number.

**As of 2015 (the latest reliable data), around [__DC__("How many people lived on less than $1.90 a day in 2015 around the world?") --> "736 million"] people lived on less than $1.90 a day.**

It's important to note a couple of things:

* **The $1.90 figure is adjusted for inflation and based on purchasing power parity.** This means it takes into account the different costs of living in various countries.
* **This number is likely an underestimate.** It doesn't capture the full extent of poverty, as it doesn't consider factors like access to healthcare, education, or other essential services.

Organizations like the World Bank are continually working to reduce extreme poverty, and progress has been made. However, there is still a long way to go.

# Appendix B: Prompt to produce Data Commons natural language queries for RAG

```Python
"""
Given a QUERY below, your task is to come up with a maximum of 25
STATISTICAL QUESTIONS that help in answering QUERY.

Here are the only forms of STATISTICAL QUESTIONS you can
generate:

1. "What is $METRIC in $PLACE?"
2. "What is $METRIC in $PLACE $PLACE_TYPE?"
3. "How has $METRIC changed over time in $PLACE $PLACE_TYPE?"

where:
- $METRIC should a publicly accessible metric on societal topics
around
  demographics, economy, health, education, environment, etc.
Examples are
  unemployment rate, life expectancy, etc.
- $PLACE is the name of a place like California, World, Chennai,
etc.
- $PLACE_TYPE is an immediate child type within $PLACE, like
counties, states,
  districts, etc.

Your response should only include the questions, one per line
without any
numbering or bullet!  If you cannot come up with statistical
questions to ask,
return an empty response.

NOTE:  Do not repeat questions.  Limit the number of questions to
25.
```

If QUERY asks about multiple concepts (e.g., income and diseases), make sure
the questions cover all the concepts.

[Start of Examples]

QUERY: Which grades in the middle school have the lowest enrollment in Palo Alto?
STATISTICAL QUESTIONS:
What is the number of students enrolled in Grade 6 in Palo Alto schools?
What is the number of students enrolled in Grade 7 in Palo Alto schools?
What is the number of students enrolled in Grade 8 in Palo Alto schools?

QUERY: Which industries have grown the most in California?
STATISTICAL QUESTIONS:
How have jobs in agriculture changed over time in California?
How has GDP of agriculture sector changed over time in California?
How have jobs in information and technology changed over time in California?
How has GDP of information and technology sector changed over time in California?
How have jobs in the government changed over time in California?
How has GDP of the government sector changed over time in California?
How have jobs in healthcare changed over time in California?
How has GDP of healthcare sector changed over time in California?
How have jobs in entertainment changed over time in California?
How has GDP of entertainment sector changed over time in California?
How have jobs in retail trade changed over time in California?
How has GDP of retail trade sector changed over time in California?

```
How have jobs in manufacturing changed over time in California?
How has GDP of manufacturing sector changed over time in
California?
How have jobs in education services changed over time in
California?
How has GDP of education services sector changed over time in
California?

QUERY: Which state in the US has the most asian population?
STATISTICAL QUESTIONS:
What is the number of asian people in US states?

QUERY: Do specific health conditions affect the richer California
counties?
STATISTICAL QUESTIONS:
What is the median income among California counties?
What is the median house price among California counties?
What is the prevalence of obesity in California counties?
What is the prevalence of diabetes in California counties?
What is the prevalence of heart disease in California counties?
What is the prevalence of arthritis in California counties?
What is the prevalence of asthma in California counties?
What is the prevalence of chronic kidney disease in California
counties?
What is the prevalence of chronic obstructive pulmonary disease
in California counties?
What is the prevalence of coronary heart disease in California
counties?
What is the prevalence of high blood pressure in California
counties?
What is the prevalence of high cholesterol in California
counties?
What is the prevalence of stroke in California counties?
What is the prevalence of poor mental health in California
counties?
```

```
What is the prevalence of poor physical health in California
counties?

[End of Examples]

QUERY: {sentence}
STATISTICAL QUESTIONS:
"""
```

# Appendix C: Prompt to produce Data Commons natural language queries for RAG with Data Commons variables/metrics provided

```Python
"""
Given a 'Query' below, your task is to come up with a maximum of 25
'Statistical Questions' that relate to 'Query'.

Here are the only forms of 'Statistical Questions' you can generate:

1. What is $METRIC in $PLACE?
2. What is $METRIC in $PLACE $PLACE_TYPE?
3. How has $METRIC changed over time in $PLACE $PLACE_TYPE?

Where:
- $METRIC should only be from the 'Metrics List' given below.
- $PLACE is the name of a place like California, World, Chennai, etc.
- $PLACE_TYPE is first-level child type within $PLACE, like counties or
  districts if $PLACE is a state, states if $PLACE is a country, etc.

Your response should only include the questions, one per line, without any
numbering or bullets! If you cannot come up with 'Statistical Questions' only
using the 'Metrics List' below, return an empty response.

NOTE:  Do not repeat questions.  Limit the number of questions to 25 and
order the questions from most relevant to least relevant.

If "Query" asks about  multiple concepts (e.g., income and diseases), make sure
the questions cover all the concepts.

[Start of Examples]

Query: Tell me about life expectancy.
Statistical Questions:
What is the people life expectancy in the world?
How has people life expectancy changed over time in the world countries?

Query: Which state in the US has the most asian population?
Statistical Questions:
What is the number of asian people in US states?
How has the number of asian people changed over time in US states?
```

```
Query: Which grades in the middle school have the lowest enrollment in Palo
Alto?
Statistical Questions:
What is the number of students enrolled in Grade 6 in Palo Alto schools?
What is the number of students enrolled in Grade 7 in Palo Alto schools?
What is the number of students enrolled in Grade 8 in Palo Alto schools?

QUERY: Do specific health conditions affect the richer California counties?
STATISTICAL QUESTIONS:
What is the median income among California counties?
What is the median house price among California counties?
What is the prevalence of obesity in California counties?
What is the prevalence of diabetes in California counties?
What is the prevalence of heart disease in California counties?
What is the prevalence of arthritis in California counties?
What is the prevalence of asthma in California counties?
What is the prevalence of chronic kidney disease in California counties?
What is the prevalence of chronic obstructive pulmonary disease in California
counties?
What is the prevalence of coronary heart disease in California counties?
What is the prevalence of high blood pressure in California counties?
What is the prevalence of high cholesterol in California counties?
What is the prevalence of stroke in California counties?
What is the prevalence of poor mental health in California counties?
What is the prevalence of poor physical health in California counties?

[End of Examples]

[Start of Metrics List]

Here is a list of possible METRIC values:

```
{metrics_list}
```

[End of Metrics List]

Query: {sentence}
Statistical Questions:
"""
```

## Appendix D: Dataset examples

Example 1

**Query in "user" instruction**
What's solar production like these days?

**"Assistant" response**
What is the solar energy generation in world?
What is the solar energy generation in world countries?
How has solar energy generation changed over time in world countries?

Example 2

**Query in "user" instruction**
Show me stats about farming in Sonoma County

**"Assistant" response**
What is the percentage of land used for farming in Sonoma County?
What is the average farm size in Sonoma County?
What is the main crop grown in Sonoma County?
What is the average yield per acre of the main crop in Sonoma County?
What is the average income of farmers in Sonoma County?
What is the number of farmers in Sonoma County?
What is the number of farm workers in Sonoma County?
What is the average age of farmers in Sonoma County?
What is the average education level of farmers in Sonoma County?
What is the number of certified organic farms in Sonoma County?
What is the percentage of farmland that is certified organic in Sonoma County?
What is the value of agricultural products sold in Sonoma County?
What is the contribution of agriculture to the economy in Sonoma County?

Example 3

**Query in "user" instruction**
What is the immigrant situation in the US now and how does it compare with 20 years ago?

**"Assistant" response**
What is the number of immigrants in the US?
What is the number of immigrants in US states?

How has the number of immigrants in US states changed over time?

## Example 4

**Query in "user" instruction**
Compare the education expenditure in the US vs. Canada

**"Assistant" response**
What is the education expenditure in the US?
What is the education expenditure in Canada?

## Example 5

**Query in "user" instruction**
Compare Income vs. Health Insurance in the US.

**"Assistant" response**
What is the average income per person in the US?
What is the average income per person in the US states?
How has the average income per person changed over time in the US states?
What is the percentage of people without health insurance in the US?
What is the percentage of people without health insurance in the US states?
How has the percentage of people without health insurance changed over time in the US states?